\documentclass[runningheads]{llncs}

\usepackage[mobile]{eccv}

\usepackage{eccvabbrv}

\usepackage{graphicx}
\usepackage{booktabs}

\usepackage[accsupp]{axessibility}  

\usepackage[pagebackref,breaklinks,colorlinks,citecolor=eccvblue]{hyperref}

\usepackage{orcidlink}

\usepackage{pifont}
\newcommand{\cmark}{\ding{51}}%
\newcommand{\xmark}{\ding{55}}%

\usepackage{amsmath,amsfonts,bm}

\def\eqref#1{equation~\ref{#1}}

\def\1{\bm{1}}

\DeclareMathAlphabet{\mathsfit}{\encodingdefault}{\sfdefault}{m}{sl}
\SetMathAlphabet{\mathsfit}{bold}{\encodingdefault}{\sfdefault}{bx}{n}

\definecolor{mycyan}{RGB}{212, 239, 251}
\usepackage{colortbl}
\usepackage{xcolor}
\definecolor{mygray}{gray}{.9}
\definecolor{goldenrod}{RGB}{245,245,220}
\newlength\savewidth
\newcolumntype{a}{>{\columncolor{mygray}}c}
\usepackage{fontawesome5}
\usepackage{booktabs}
\usepackage{makecell}
\usepackage{fontawesome5}
\usepackage{lipsum}
\usepackage{comment}
\usepackage{multirow}
\usepackage{wrapfig}
\usepackage{algorithm}
\usepackage{algorithmic}
\usepackage{xfrac}  

\usepackage{graphicx}
\usepackage{color}

\definecolor{darkgreen}{rgb}{0,0.7,0}

\definecolor{mygraytext}{gray}{.5}

\def\eg{\emph{e.g.}}

\begin{document}

\title{EfficientVMamba: Atrous Selective Scan for\\Light Weight Visual Mamba} 

\titlerunning{EfficientVMamba}

\author{Xiaohuan Pei\thanks{Equal contributions.} \and Tao Huang$^*$ \and Chang Xu}

\authorrunning{X.~Pei et al.}

\institute{School of Computer Science, Faculty of Engineering, The University of Sydney \email{\{xpei8318,thua7590\}@uni.sydney.edu.au}, \email{c.xu@sydney.edu.au}
}

\maketitle

\begin{abstract}
Prior efforts in light-weight model development mainly centered on CNN and Transformer-based designs yet faced persistent challenges. CNNs adept at local feature extraction compromise resolution while Transformers offer global reach but escalate computational demands $\mathcal{O}(N^2)$. This ongoing trade-off between accuracy and efficiency remains a significant hurdle. Recently, state space models (SSMs), such as Mamba, have shown outstanding performance and competitiveness in various tasks such as language modeling and computer vision, while reducing the time complexity of global information extraction to $\mathcal{O}(N)$. Inspired by this, this work proposes to explore the potential of visual state space models in light-weight model design and introduce a novel efficient model variant dubbed EfficientVMamba. Concretely, our EfficientVMamba integrates a atrous-based selective scan approach by efficient skip sampling, constituting building blocks designed to harness both global and local representational features. Additionally, we investigate the integration between SSM blocks and convolutions, and introduce an efficient visual state space block combined with an additional convolution branch, which further elevate the model performance. Experimental results show that, EfficientVMamba scales down the computational complexity while yields competitive results across a variety of vision tasks. For example, our EfficientVMamba-S with $1.3$G FLOPs improves Vim-Ti with $1.5$G FLOPs by a large margin of $5.6\%$ accuracy on ImageNet. Code is available at: \url{https://github.com/TerryPei/EfficientVMamba}.

\end{abstract}

\section{Introduction}
\label{sec:intro}

Convolutional networks, exemplified by models such as ResNet\cite{he2016deep}, Inception \cite{szegedy2016rethinking, szegedy2017inception}, and EfficientNet\cite{tan2019efficientnet}, 
and 
Transformer-based networks, such as Swin-Transformer \cite{liu2021swin}, Beit \cite{bao2021beit}, and Resformer \cite{zamir2022restormer}   
have been extensively applied to visual tasks including image classification, detection, and segmentation, achieving remarkable results. Recently, Mamba\cite{gu2023mamba}, a network based on state-space models (SSMs) \cite{gu2021combining, gu2021efficiently, gupta2022diagonal, li2022makes, orvieto2023resurrecting}, has demonstrated competitive performance to Transformers \cite{vaswani2017attention} in sequence modeling tasks such as language modeling. Inspired by this, some works \cite{zhu2024vision,liu2024vmamba,ruan2024vm,xing2024segmamba,liu2024swin} are pioneering in introducing SSMs into vision tasks. Among these methods, Vmamba\cite{liu2024vmamba} stands out by introducing an SS2D method to preserve 2D spatial dependencies by scanning images from multiple directions. 

\begin{wrapfigure}{r}{0.5\textwidth}
  \vspace{-2em}
  \setlength{\abovecaptionskip}{0.00cm}
  \setlength{\belowcaptionskip}{0.00cm}
  \begin{center}
    \includegraphics[width=0.49\textwidth]{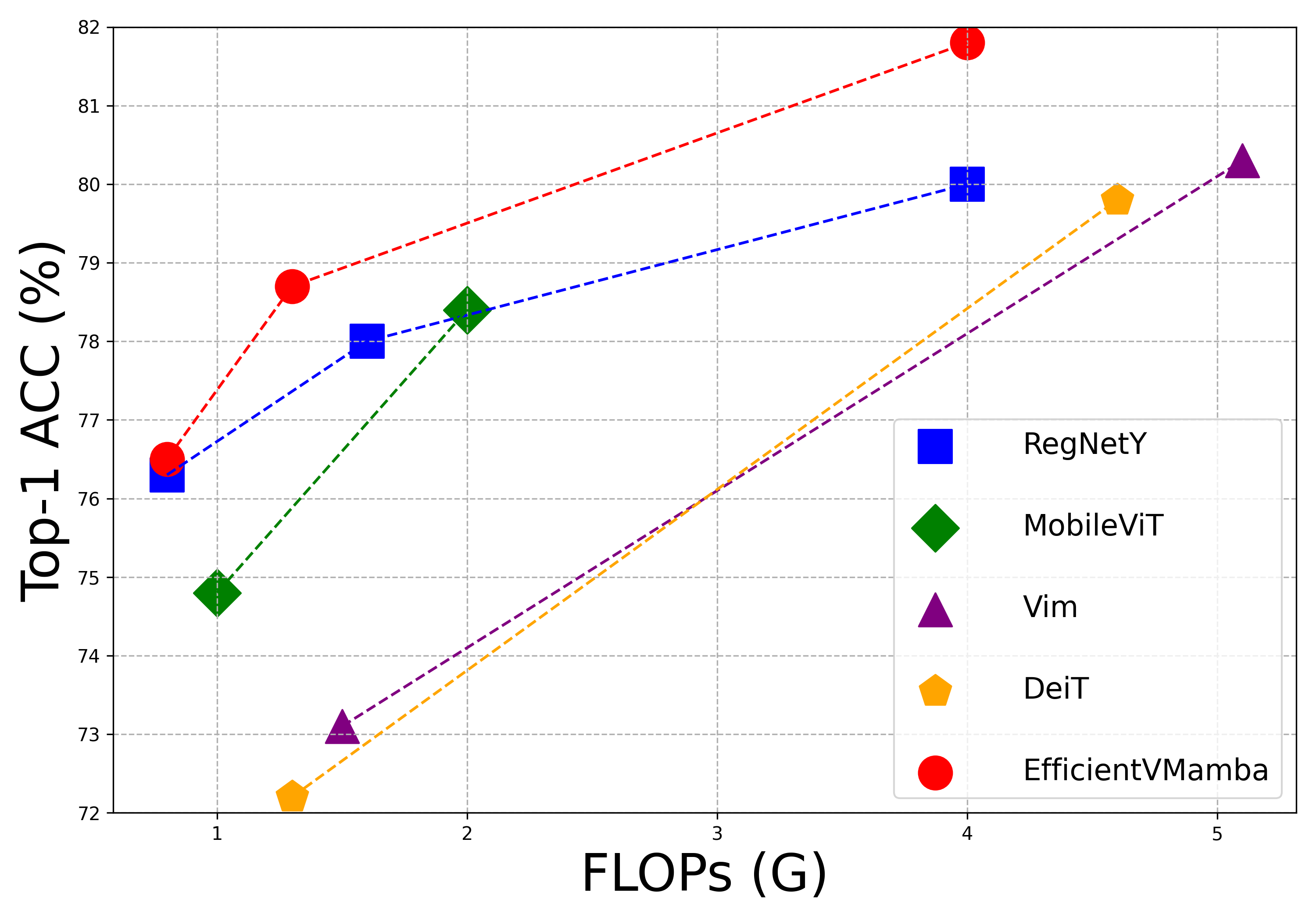}
  \end{center}
  \caption{Lightweight Model Performance Comparison on ImageNet. EfficientVMamba outperforms previous work across various model variants in terms of both accuracy and computational complexity.}
  \vspace{-0.1in}
  \label{fig:flops}
\end{wrapfigure}
However, the impressive performance achieved by these various architectures usually comes from the scaling up of model sizes, making a critical challenge in applying them on resource-constrained devices.
In pursue of light-weight models, many studies have been conducted to reduce the resource consumption of vision models while keeping a competitive performance. Early works on efficient CNNs mainly focus on narrowing the original convolutional block with efficient group convolutions \cite{chollet2017xception,howard2017mobilenets,sandler2018mobilenetv2}, light skipping connections \cite{ma2018shufflenet, han2020ghostnet}, \textit{e.t.c}. While recently, due to the remarkable success of including the global representation ability of Transformers into vision tasks, some works are proposed to reduce the computation complexity of ViTs \cite{liu2021swin,tu2022maxvit,wang2021pyramid,huang2022lightvit} and fuse ViTs with CNNs in light-weight models \cite{mehta2021mobilevit, li2022efficientformer, tu2022maxvit}.
However, the lightening of ViTs are usually obtained with the lost of global capture capability in self-attention. Due to the $\mathcal{O}(N)$ time complexity of global self-attention, its computation and memory costs increase dramastically at large resolutions. As a result, existing efficient ViT methods have to perform local self-attentions within partitioned windows \cite{liu2021swin,tu2022maxvit, huang2022lightvit}, or only conduct global self-attentions in deeper stages with low resolutions \cite{mehta2021mobilevit, li2022efficientformer}. The embarrassing trade-off and rollback of ViTs to CNNs hinders the ability of improving the light-weight models further.

\begin{figure}[t]
    \centering
    \includegraphics[width=\linewidth]{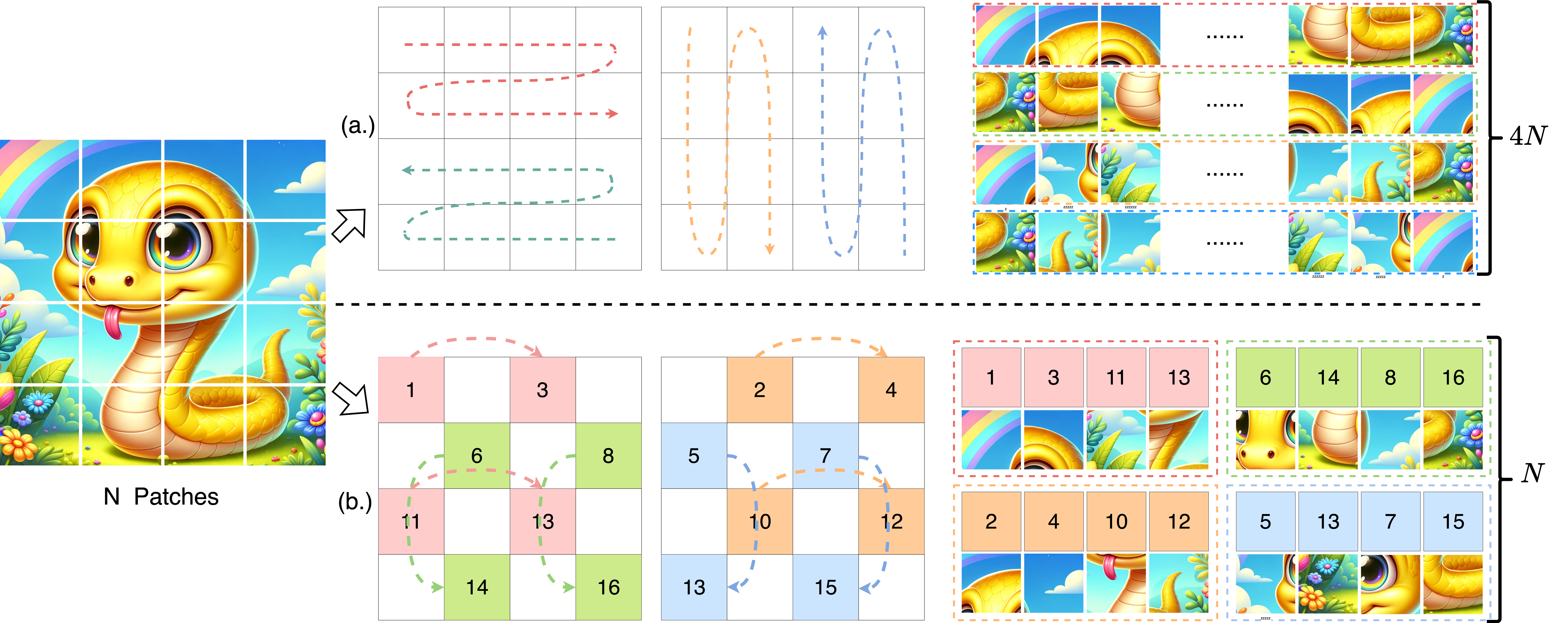}
    \caption{Illustration of efficient 2D scan methods (ES2D). 
    (a.) Vmamba \cite{liu2024vmamba} employs SS2D method in vision tasks, traversing entire row or column axes, which incurs heavy computational resources. (b.) We present an efficient 2D scanning method, ES2D, which organizes patches by omitting sampling steps, and then proceeds with an intra-group traversal (with a skipping step of 2 in the Figure). The proposed scan approach reduces computational demands ($4N \rightarrow N$) while preserving global feature maps (\eg Each group contains eye-related patches.)
    }
    \label{fig:scan}
    \vspace{-0.5em}
\end{figure}

In this paper, recalling the previously metioned linear scaling complexity in SSMs, we are inspired to obtain efficient global capture ability in light-weight vision models by involving SSMs into model design. Its outstanding performance is demonstrated in Figure \ref{fig:flops}.  We achieve this by first introducing a skip-sampling mechanism, which reduces the number of tokens that need to be scanned in the spatial dimension, and saves multiple times of computation cost in sequence modeling of SSMs while keeping the global receptive field among tokens, as illustrated in Figure \ref{fig:scan}. On the other hand, acknowledging that the convolutions provide a more efficient way for feature extraction in the case when only local representations suffice, we introduce a convolution branch in supplement of the original global SSM branch, and perform feature fusion of them through the channel attention module, SE \cite{hu2018squeeze}. Finally, for an optimal allocation of capabilities of various block types, we construct our network with global SSM blocks in the shallow and high-resolution layers, while adopting efficient convolution blocks (MobileNetV2 blocks \cite{sandler2018mobilenetv2}) in the deeper layers. The final network, achieving efficient SSM computation and efficient integration of convolutions, has showcased significant improvements compared to previous CNN and ViT based light-weight models through our experiments on image classification, object detection, and semantic segmentation tasks.

In summary, the contributions of this paper are as follows.
\begin{enumerate}
    \item We propose an atrous-based selective scanning strategy, which is realized through a novel skip sampling and regrouping patched in the spatial respective field. The strategy refines the building blocks to efficiently extract global dependencies while reducing computation complexity ($\mathcal{O}(N) \rightarrow \mathcal{O}(N/p^2)$) with step $p$). 
    \item We introduce a dual-pathway module that combines our efficient scanning strategy for global feature capture and a convolution branch for efficient local feature extraction, along with a channel attention module to balance the integration of both global and local features. Besides, we propose a better allocation of SSM and CNN blocks by promoting SSMs in early stages with high resolutions for better global capture, while adopting CNNs in low resolutions for better efficiency.
    \item We conduct extensive experiments on image classification, object detection, and semantic segmentation tasks. The results and illustration shown in Figure \ref{fig:flops} demonstrate that, our EfficientVMamba effectively reduces the FLOPs of the models while achieving significant performance improvements compared to existing light-weight models.
\end{enumerate}

\section{Related Work}
\label{sec:related}

\subsection{Light-weight Vision Models}
\label{subsec:Light-weight Vision Models}
In recent years, the realm of vision tasks has been predominantly governed by Convolutional Neural Networks (CNNs) and Visual Transformer (ViT) architectures. The focus on making these architectures lightweight to enhance efficiency has emerged as a pragmatic and promising direction in research. For CNNs, notable advancements have been made in improving image classification accuracy, as evidenced by the development of influential architectures like ResNet\cite{he2016deep}, RegNet\cite{schneider2017regnet}, and DenseNet\cite{iandola2014densenet}. These advancements have set new benchmarks in accuracy but also introduced a need for lightweight architectures\cite{wang2018learning,wang2018towards}. This need has been addressed through various factorization-based methods, making CNNs more mobile-friendly. For instance, separable convolutions introduced by Xception have been instrumental in this regard, leading to the development of state-of-the-art lightweight CNNs, such as MobileNets\cite{howard2017mobilenets}, ShuffleNetv2\cite{ma2018shufflenet}, ESPNetv2\cite{mehta2019espnetv2}, MixConv\cite{tan2019mixconv}, MNASNet\cite{tan2019mnasnet}, and GhostNets\cite{han2022ghostnets}. These models are not only versatile but also relatively simpler to train.
Following CNNs, Transformers have gained significant traction in various vision tasks, such as image classification, object detection, and autonomous driving, rapidly becoming the mainstream approach. The lightweight versions of Transformers have been achieved through diverse methods. On the training front, sophisticated data augmentation strategies and techniques like Mixup\cite{zhang2017mixup}, CutMix\cite{yun2019cutmix}, and RandAugment\cite{cubuk2020randaugment} have been employed, as seen in models like CaiT \cite{touvron2021going} and DeiT-III \cite{touvron2022deit}, which demonstrate exceptional performance without the need for large proprietary datasets. From the architectural design perspective, efforts have been concentrated on optimizing self-attention input resolution and devising attention mechanisms that incur lower computational costs. Innovations like PVT-v1 \cite{wang2021pyramid}'s emulation of CNN's feature map pyramid, Swin-T \cite{liu2021swin} and LightViT\cite{huang2022lightvit}'s hierarchical feature map and shifted-window mechanisms, and the introduction of (multi-scale) deformable attention modules in Deformable DETR \cite{zhu2020deformable} exemplify these advancements. There is also NAS for ViTs \cite{su2022vitas}.

\subsection{State Space Models}
\label{subsec:State Space Models}
The State Space Model (SSM) \cite{gu2021combining, gu2021efficiently, gupta2022diagonal, li2022makes, orvieto2023resurrecting} is a family architecture encapsulates a sequence-to-sequence transformation has the potential to handle tokens with long dependencies, but it is challenging to train due to its high computational and memory usage. Nevertheless, recent works \cite{gu2021efficiently, gupta2022diagonal, gu2022parameterization, gu2023mamba, smith2022simplified} have enabled deep State Space Models to become progressively more competitive with CNN and Transformer.
In particular, S4 \cite{gu2021efficiently} employs a Normal Plus Low-Rank (NPLR) representation to efficiently compute the convolution kernel by leveraging the Woodbury identity for matrix inversion. And then Mamba \cite{gu2023mamba} enhances SSMs with input-specific parameterization and a scalable, hardware-optimized algorithm, achieving simpler design and superior efficiency in processing long sequences for language and genomics.
Following success of SSM, there has been a surge in applying the framework to computer vision tasks. S4ND \cite{nguyen2022s4nd} first introduce the SSM blocks into vision tasks, facilitating the modeling of visual data across 1D, 2D, and 3D as continuous signals. 
Vmamba \cite{liu2024vmamba} pioneers a mamba-based vision backbone  a cross-scan module to address the direction-sensitivity issue arising from the differences between 1D sequences and multi-channels images.
Similarly, Vim \cite{zhu2024vision} introduces an efficient state space model for vision tasks by leveraging bidirectional state space modeling for data-dependent global visual context without image-specific biases.
The impressive performance of the Mamba backbone in various vision tasks has inspired a wave of research \cite{ruan2024vm, ruan2024vm, liu2024swin, wang2024graph, behrouz2024graph} focusing on adapting Mamba-based models for specialized vision applications. 
Recent works like Vm-unet \cite{ruan2024vm}, U-Mamba \cite{liu2024swin}, and SegMamba \cite{xing2024segmamba} have adapted Mamba-based backbones for medical image segmentation, integrating unique features such as a U-shaped architecture in Vm-unet, an encoder-decoder framework in U-Mamba, and whole volume feature modeling in SegMamba.
In the domain of graph representation, GraphMamba \cite{wang2024graph} integrates Graph Guided Message Passing (GMB) with Message Passing Neural Networks (MPNN) within the Graph GPS architecture, which enhances the training and contextual filtration for graph embeddings. Furthermore, GMNs \cite{behrouz2024graph} present a comprehensive framework that encompasses tokenization, optional positional or structural encoding, localized encoding, sequencing of tokens, and utilizes a series of bidirectional Mamba layers for processing graphs.

\section{Preliminaries}
\subsection{State Space Models (S4)}
State Space Models (SSMs) are a general family of sequence model used in deep learning that are influenced by systems capable of mapping one-dimensional sequences in a continuous manner. These models transform input $D$-dimensional sequence $x(t) \in \mathbb{R}^{L \times D}$   into output sequence $y(t) \in \mathbb{R}^{L \times D}$ by utilizing a learnable latent state $h(t) \in \mathbb{R}^{N \times D}$ that is not directly observable. The mapping process could be denoted as:
\begin{align}\label{eq:ode}
    \begin{split}
        h'(t) &= \bm{A}h(t) + \bm{B}x(t),\\
        y(t) &= \bm{C}h(t), 
    \end{split}
\end{align}
where $\bm{A}\in\mathbb{R}^{N\times N}$, $\bm{B}\in\mathbb{R}^{D\times N}$ and $\bm{C}\in\mathbb{R}^{D \times N}$. 

\textbf{Discretization.} Discretization aims to convert the continuous differential equations into discrete functions, aligning the model to the input signal's sampling frequency for more efficient computation \cite{gu2021combining}. Following the work \cite{gupta2022diagonal}, the 
continuous parameters ($\bm{A}$, $\bm{B}$) can be discretized by zero-order hold rule with a given sample timescale $\Delta \in \mathbb{R}^{D}$:

\begin{equation} \label{eq:discretization}
\begin{aligned}
&\bar{\bm{A}} = e^{\Delta \bm{A}}, \\
&\bar{\bm{B}} = (e^{\Delta \bm{A}} - \bm{I})\bm{A}^{-1}\bm{B}, \\
&\bar{\bm{C}} = \bm{C}, \\
&\bar{\bm{B}} \approx (\Delta \bm{A})(\Delta \bm{A})^{-1}\bm{A}\bm{B} = \Delta \bm{B},\\
&h(t) = \bar{\bm{A}}h(t-1) + \bar{\bm{B}}x(t), \\
&y(t) = \bar{\bm{C}}h(t), 
\end{aligned}
\end{equation}
where $\bm{\bar{A}}\in\mathbb{R}^{N\times N}$, $\bm{\bar{B}} \in \mathbb{R}^{D\times N}$ and $\bm{\bar{C}}\in\mathbb{R}^{D \times N}$. 

To simplify calculations, the repeated application of Equation \ref{eq:discretization} can be efficiently performed simultaneously using a global convolution approach.
\begin{align}
    \begin{split}
        \bm{y} &= \bm{x} \circledast \bm{\overline{K}} \\
        \text{with} \quad \bm{\overline{K}} &= (\bm{C}\bm{\overline{B}},\bm{C}\overline{\bm{A}\bm{B}}, ..., \bm{C}\bm{\overline{A}}^{L-1}\bm{\overline{B}}),
    \end{split}
\end{align}
where $\circledast$ denotes convolution operation, and $\bm{\overline{K}} \in \mathbb{R}^{L}$ is the SSM kernel.

\subsection{Selective State Space Models (S6)}
Mamba \cite{gu2023mamba} improves the performance of SSM by introducing Selective State Space Models (S6), allowing the continuous parameters to vary with the input enhances selective information processing across sequences, which extend the discretization process by selection mechanism:
\begin{equation}
\begin{aligned}
\bar{\bm{B}} &= s_{\bm{B}}(x), \\
\bar{\bm{C}} &= s_{\bm{C}}(x), \\
\Delta &= \tau_{\bm{A}}(\text{Parameter} + s_{\bm{A}}(x)), 
\end{aligned}
\end{equation}
where $ s_{\bm{B}}(x) $ and $ s_{\bm{C}}(x) $ are linear functions that project input $ x $ into an N-dimensional space, while $ s_{\bm{A}}(x) $ broadens a $D$-dimensional linear projection to the necessary dimensions.
In terms of visual tasks, VMamba proposed the 2D Selective Scan (SS2D) \cite{liu2024vmamba}, which maintains the integrity of 2D image structures by scanning four directed feature sequences. Each sequence is processed independently within an S6 block and then being combined to form a comprehensive 2D feature map.

\section{Method}
\label{sec:Method}
\begin{figure}[t]
    \centering
    \includegraphics[width=\linewidth]{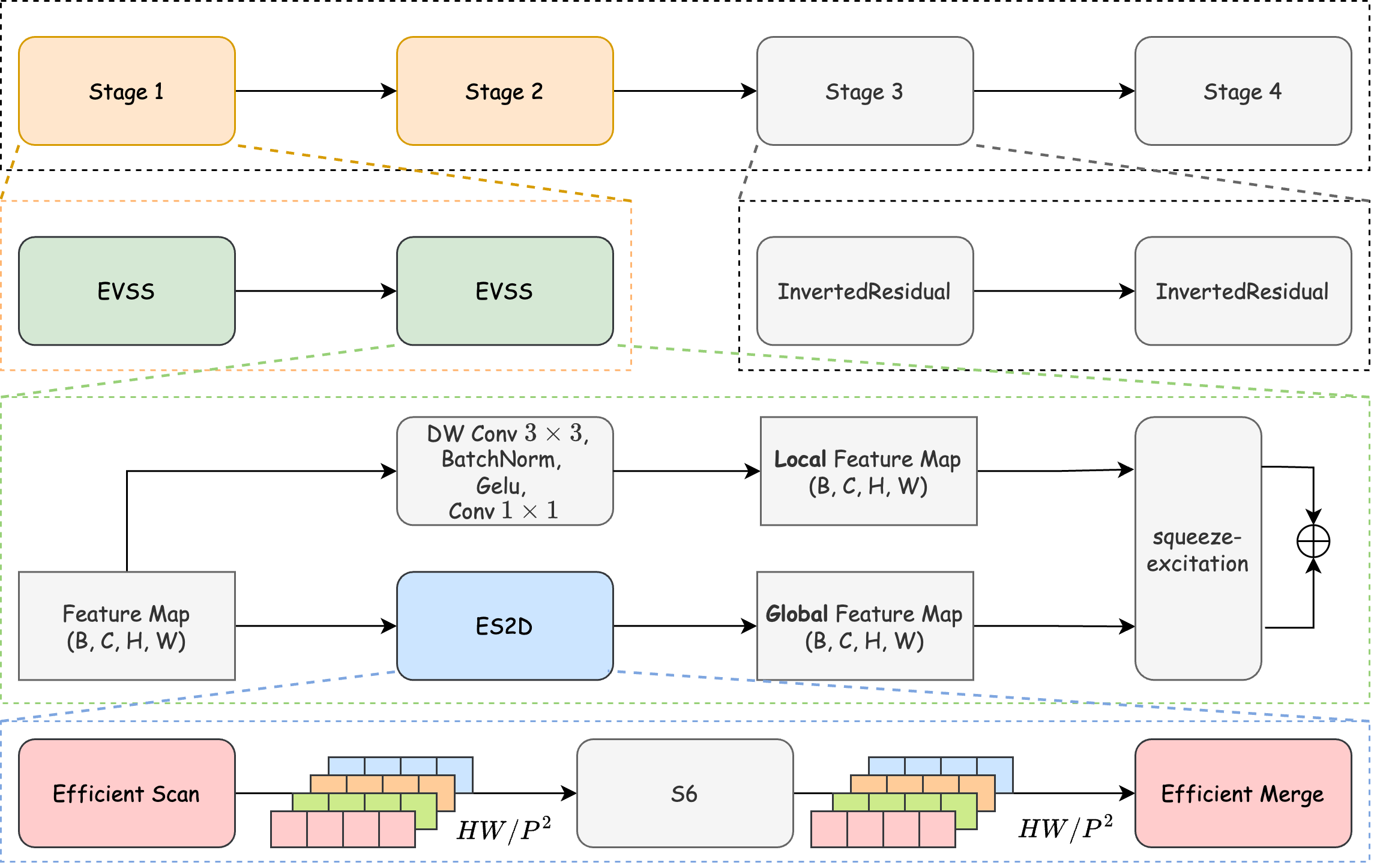}
    \caption{Architecture overview of EfficientVMamba. We hightlight our contributions with corresponding colors in the Figure. (1) \textcolor{blue}{ES2D} \ref{sec:ES2D}: Atrous-based selective scanning strategy via skip sampling and regrouping in the spatial space.  (2) \textcolor{green}{EVSS} \ref{sec:EVSS}: The EVSS block merges global and local feature extraction with modified ES2D and convolutional approaches enhanced by Squeeze-Excitation blocks for refined dual-pathway feature representation. \textcolor{orange}{Inverted Fusion} \ref{sec:Inverted}: Inverted Fusion places local-representation modules in deep layers, deviating from traditional designs by utilizing EVSS blocks early for global representation and inverted residual blocks later for local feature extraction.
    }
    \label{fig:arch}
    \vspace{-1.5em}
\end{figure}

In order to design light-weight models that are friendly to resource-limited devices, we propose EfficientVMamaba, which is summarized in Figure \ref{fig:arch}. We introduce an efficient selective scan approach to reduce the computational complexity in Section \ref{sec:ES2D}, and build a block considering both global and local feature extraction with the integration of SSMs and CNNs in Section \ref{sec:EVSS}. Regarding the design of architecture, Section \ref{sec:model variants} then offers an in-depth look at various architectural variations tailored to different model sizes.

\subsection{
Efficient 2D Scanning (ES2D)}
\label{sec:ES2D}
In deep neural networks, downsampling via pooling or strided convolution is employed to broaden the receptive field with a lower computational cost; however, this comes at the expense of spatial resolution. Previous work \cite{yu2015multi, tu2022maxvit} demonstrate apply atrous-based strategy benefits broadening the receptive field without sacrificing resolution. 
Inspired by this observation and aiming to alleviate and light the computational complexity of selective scanning,
we propose an efficient 2D scanning (ES2D) method to scale down the visual selective scan  block (SS2D) via skipping sampling for each patches on the feature map. Given a input feature map $\bm{X} \in \mathbb{R}^{C \times H \times W}$, instead of cross-scan whole patches, we skip scan patches with a step size $p$ and partition into selected 
spatial dimensional features $\{\bm{O}_i\}_{i=1}^{4} $:
\begin{equation} \label{eq:scan}
\begin{aligned}
&\bm{O}_{i} \xleftarrow{\text{scan}} \bm{X}[:, m::p, n::p], \\
&\{\bm{\tilde{O}}_i\}_{i=1}^{4} \leftarrow \text{SS2D}(\{\bm{O}_i\}_{i=1}^{4}), \\
&\bm{Y}[:, m::p, n::p] \xleftarrow{\text{merge}} \bm{\tilde{O}}_{i}, \\
\end{aligned}
\end{equation}
\begin{equation*}
    \text{with} \quad  (m, n) = ( \left\lfloor \frac{1}{2} + \frac{1}{2} \sin\left(\frac{\pi}{2} (i - 2)\right) \right\rfloor, \left\lfloor \frac{1}{2} + \frac{1}{2} \cos\left(\frac{\pi}{2} (i - 2)\right) \right\rfloor ),
\end{equation*}
where  $\bm{O}_i, \bm{\tilde{O}}_i \in \mathbb{R}^{C \times \frac{H}{p} \times \frac{W}{p}}$ and the operation $[:, m::p, n::p]$ represents slicing the matrix for each channel, starting at $m$ on height ($H$) and $n$ on width ($W$), skipping every $p$ steps.
The process decompose the fully scanning method into both local and global sparse forms. Skip sampling for local receptive fields reduces computational complexity by selectively scanning smaller patches of the feature map. With a step size $p$, we sample the $(C, H/p, W/p)$ patches at intervals of $p$, compared to $(C, H, W)$ in the SS2D, decreasing the number of tokens processed from $N$ to $\frac{N}{p^2}$ for each scan and merge operation, which improves feature extraction efficiency.
Re-grouping for global spatial feature maps in ES2D involves combining the processed patches to reconstruct the global structure of the feature map. This integration captures broader contextual information, balancing local detail and global context in feature extraction.
Accordingly, our design is intended to streamline the scanning and merging modules while maintaining the essential benefit of global integration in the state-space architecture, with the aim of ensuring that the feature extraction remains comprehensive on the spatial axis.

\subsection{Efficient Visual State Space Block (EVSS)} %
\label{sec:EVSS}
Based on the efficient selected scan approach, we introduce the Efficient Visual State Space (EVSS) block, which is designed to synergistically merge global and local feature representations while maintaining computational efficiency. It leverages a SqueezeEdit-modified ES2D for global information capture and a convolutional branch tailored to extract critical local features, with both branches undergoing a subsequent Squeeze-Excitation (SE) block \cite{hu2018squeeze}.
The ES2D module aims to efficiently abstract global contextual information by implementing an intelligent skipping mechanism presented in \ref{sec:ES2D}. It selectively scans the map with a step size \( p \), reducing redundancy without sacrificing the representational quality of the global context in the resultant spatial dimensional features. 
Parallel to this, empirical evidence concurs that convolutional operations offer a more proficient approach to feature extraction, particularly in scenarios where local representations are adequate. We add the convolutional branch concentrates on discerning fine-grained local details through a \( 3 \times 3 \) convolution of stride 1. 
The subsequent SE block adaptively recalibrates the features, allowing the network to auto re-balanced the local and global respective field on the feature map.

The outputs of the respective SE blocks are combined via element-wise summation to construct the EVSS's output and the dual pathway could be denoted as:
\begin{equation} \label{eq:EVSS}
    \bm{X}^{l+1} = \text{SE}(\text{ES2D}(\bm{X}^l)) + \text{SE}(\text{Conv}(\bm{X}^l)),
\end{equation}
where $\bm{X}^l$ represent the feature map of the $l$-layer and $\text{SE}(\cdot)$ is the Squeeze-Excitation operation.
With each pathway utilizing a SE block, the EVSS ensures that the respective features of global and local information are dynamically re-balanced to emphasize the most salient features.
This fusion aims to preserve the integrity of both the expansive global perspective and the intricate local specifics, facilitating a comprehensive feature representation.

\subsection{Inverted Insertion of EfficientNet Blocks}
\label{sec:Inverted}
As a well-established consensus, the computational efficiency of convolutional operations is more efficient than that of the global-based block such as Transformer. 
Prior light-weight work efforts have predominantly employed computation-efficient convolutions in the former stages to scale down the token numbers to reduce computational complexity, subsequently integrating global-based blocks (\eg, Transformer with the computational complexity of $\mathcal{O}(N^2)$) to capture global context in the latter stages. For example, MobileViT \cite{mehta2021mobilevit} adopts pure MobileNetV2 blocks in the first two downsampling stages, while only integrating self-attention operations in the latter stages at low resolutions. EfficientFormer \cite{li2022efficientformer} introduces two types of base blocks, the convolution-based blocks with local pooling are used in the first three stages, and the transformer-like self-attention blocks are only leveraged in the last stage.

However, the observation is contrast on the Mamba-based block. In the SSM framework, the computational complexity for global representation is  $\mathcal{O}(N)$, indicating that placing local representation modules at either the front or the back of the stage could be reasonable. Through empirical observation in Table \ref{tab:ablation2}, we found positioning these local-representation modules towards the latter layers of the stage yields better results. This discovery significantly deviates from the design principles of previous CNN-based and Transformer-based lightweight models, thereby we call it \texttt{inverted} insertion. Consequently, our designed $L$ stages architecture is an inverted insertion of EfficientNet Blocks (MobileNetV2 blocks with SE modules), which utilizes EVSS blocks \ref{sec:EVSS} in the former two stages to capture global-representation and Inverted Residual blocks $\text{InRes}(\cdot)$ \cite{sandler2018mobilenetv2} in the subsequent stages to extract local feature maps:
\begin{equation} \label{eq:}
\bm{X}^{l+1} = 
\begin{cases}
    & \text{EVSS}(\bm{X}^l) \qquad \text{if} \quad X^{l} \in \{\text{stage} 1, \text{stage} 2 \}; \\
    & \text{InRes}(\bm{X}^l) \qquad \text{otherwise},
\end{cases} 
\end{equation}
where $\bm{X}^l$ is the feature map in the $l$-layer. The inverted insertion design of using the shortcuts directly between the bottlenecks is considerably more memory efficient \cite{sandler2018mobilenetv2}.

\begin{table}[t]
\centering
\caption{Model variants of EfficientVMamba.}
\scriptsize
\renewcommand{\arraystretch}{1.4}
\setlength{\tabcolsep}{0.4mm}
\begin{tabular}{@{}l|c|c|c|c@{}}
\toprule
Layer & Feature Size & EfficientVMamba-T & EfficientVMamba-S & EfficientVMamba-B \\ 
\midrule
Stem & 112 $\times$ 112 & \#Dim=[48, 96, 192, 384] & \#Dim=[96, 192, 384, 768] &  \#Dim=[96, 192, 384, 768]\\ \midrule
Stage 1 & 56 $\times$ 56 & EVSS $\times$ 2 & EVSS $\times$ 2 & EVSS $\times$ 2 \\ 
Stage 3 & 14 $\times$ 14 & InRes $\times$ 4 & InRes $\times$ 4 & InRes $\times$ 9 \\
Stage 4 & 7 $\times$ 7 & InRes $\times$ 2 & InRes $\times$ 2 & InRes $\times$ 2 \\ \midrule
 & 1 $\times$ 1 & \multicolumn{3}{c}{Average pool, Fc, Softmax} \\ \midrule
\multicolumn{2}{c|}{Params. (M)} & 6  & 11  & 33 \\ 
\multicolumn{2}{c|}{FLOPs (G)} &  0.8  & 1.3  & 4.0  \\ \bottomrule
\end{tabular}
\label{tab:Model Variants}
\vspace{-0.5em}
\end{table}
\subsection{Model Variants} \label{sec:model variants}
To sufficiently demonstrate the effectiveness of our proposed model, we detail architectural variants rooted in plain structures as referenced in \cite{zhu2024vision}. These variants are designated as EfficientVMamba-T, EfficientVMamba-S, and EfficientVMamba-B, shown as Table \ref{tab:Model Variants}, corresponding to different scales of the model.
EfficientVMamba-T is the most lightweight with 6M parameters, followed by EfficientVMamba-S with 11M, and EfficientVMamba-B being the most complex with 33M. In terms of computational load, measured in FLOPs, the models exhibit a parallel increase with 0.8G for EfficientVMamba-T, 1.3G for EfficientVMamba-S, and 4.0G for EfficientVMamba-B, correlating directly with their complexity and feature size.

\section{Experiments}
To rigorously evaluate the performance of our diverse model variants, we demonstrate the results of image classification task in Section \ref{sec:ImageNet Classification}, investigate object detection performance in Section \ref{sec:Object Detection} and explore the image semantic segmentation in Section \ref{sec:Semantic Segmentation}. In section \ref{sec:ablation} We further pursued ablation study to comprehensively examine the effects of atrous selective scanning , the impact of SSM-Conv fusion blocks, and the implications of incorporating convolution blocks at different stages of the models.

\subsection{ImageNet Classification}
\begin{table}[t]
    \centering
    \setlength{\tabcolsep}{2mm}
    \renewcommand{\arraystretch}{1.1}
    \caption{Comparison of different backbones on ImageNet-1K classification.}
    \small
    \begin{tabular}{l|ccc|c}
    \toprule
    Method & Image size & Params (M) & FLOPs (G) & Top-1 ACC (\%)\\
    \midrule
    RegNetY-800M \cite{radosavovic2020designing} & $224^2$ & 6 & 0.8 & 76.3\\
    PVTv2-B0 \cite{wang2022pvt} & $224^2$ & 3 & 0.6 & 70.5\\
    MobileViT-XS \cite{mehta2021mobilevit} & $224^2$ & 2 & 1.0 & 74.8\\
    LVT \cite{yang2022lite} & $224^2$ & 6 & 0.9 & 74.8\\
    \rowcolor{mygray}
    EfficientVMamba-T & $224^2$ & 6 & 0.8 & 76.5 \\
    \midrule
    RegNetY-1.6G \cite{radosavovic2020designing} & $224^2$ & 11 & 1.6 & 78.0\\
    DeiT-Ti \cite{touvron2021training} & $224^2$ & 6 & 1.3 & 72.2\\
    MobileViT-S \cite{mehta2021mobilevit} & $224^2$ & 6 & 2.0 & 78.4\\
    Vim-Ti \cite{zhu2024vision} & $224^2$ & 7 & 1.5 & 73.1\\
    \rowcolor{mygray}
    EfficientVMamba-S & $224^2$ & 11 & 1.3 & 78.7 \\
    \midrule
    RegNetY-4G \cite{radosavovic2020designing} & $224^2$ & 21 & 4.0 & 80.0\\
    DeiT-S \cite{touvron2021training} & $224^2$ & 22 & 4.6 & 79.8\\
    Swin-T \cite{liu2021swin} & $224^2$ & 29 & 4.5 & 81.3\\
    Vim-S \cite{zhu2024vision} & $224^2$ & 26 & 5.1 & 80.3\\
    VMamba-T \cite{liu2024vmamba} & $224^2$ & 22 & 5.6 & 82.2\\
    \rowcolor{mygray}
    EfficientVMamba-B & $224^2$ & 33 & 4.0 & 81.8 \\
    \bottomrule
    \end{tabular}
    \label{tab:image classification}
    \vspace{-0.5em}
\end{table}
\label{sec:ImageNet Classification}
\textbf{Training strategies.} Following previous works \cite{touvron2021training,liu2021swin,zhu2024vision,liu2024vmamba}, we train our models for $300$ epochs with a base batch size of $1024$ and an AdamW optimizer, a cosine annealing learning rate schedule is adopted with initial value $10^{-3}$ and 20-epoch warmup. For training data augmentation, we use random cropping, AutoAugment \cite{cubuk2019autoaugment} with policy \textit{rand-m9-mstd0.5}, and random erasing of pixels with a probability of $0.25$ on each image, then a MixUp\cite{zhang2017mixup} strategy with ratio $0.2$ is adopted in each batch. An exponential moving average on model is adopted with decay rate $0.9999$.

\textbf{Tiny Models ($FLOPs (G) \in [0, 1]$)}. In the pursuit of efficiency, the results of tiny models are shown in Table \ref{tab:image classification}. EfficientVMamba-T achieves state-of-art performance with a Top-1 accuracy of 76.5\%, rivalling its counterparts that demand higher computational costs. With a modest expenditure of only $0.8$ GFLOPs, our model surpasses the PVTv2-B0 by a 6\% margin in accuracy and outperforms the MobileViT-XS by 1.7\%, all with less computational demand. 

\textbf{Small Models ($FLOPs (G) \in [1, 2]$)}.
Our model, EfficientVMamba-S, exhibits a significant improvement in accuracy, achieving a Top-1 accuracy of $78.7\%$. This represents a substantial increase over DeiT-Ti and MobileViT-S, which achieve $72.2\%$ and $78.4\%$ respectively. Notably, EfficientVMamba-S maintains this high accuracy level with computational efficiency, requiring only $1.3$ GFLOPs, which is on par with DeiT-Ti and lower than MobileViT-S's $2.0$ GFLOPs. 

\textbf{Base Models ($FLOPs (G) \in [4, 5]$)}.
EfficientVMamba-B achieves an impressive Top-1 accuracy of $81.8\%$, surpassing DeiT-S by $2\%$ and Vim-S by $1.5\%$, as indicated in the third group of the Table \ref{tab:image classification}. This base model demonstrates the feasibility of coupling a substantial parameter count of $33$ M with a modest computational demand of $4.0$ GFLOPs. In comparison, VMamba-T, with a similar parameter count of $22$ M requires a higher $5.6$ GFLOPs.

\subsection{Object Detection}
\label{sec:Object Detection}

\textbf{Training strategies.} We evaluate the efficacy of our EfficientVMamba model for object detection tasks on the MSCOCO 2017 \cite{lin2014microsoft} dataset. Our evaluation framework relies on the mmdetection library \cite{chen2019mmdetection}. For comparisons with light-weight backbones, we follow PvT \cite{wang2021pyramid} to use RetinaNet as the detector and adopt 1$\times$ training schedule. While for comparisons with larger backbones, our experiment follows the hyperparameter settings detailed in Swin \cite{liu2021swin}
We use the AdamW optimization method to refine the weights of our pre-trained networks on ImageNet-1K for durations of 12 and 36 epochs. We apply drop path rates of 0.2\% across the board for EfficientVMamba-T/S/B variants. The learning rate begins at $1e-5$ and is decreased tenfold at epochs 9 and 11. Multi-scale training and random flipping are implemented during training with a batch size of 16, adhering to standard procedures for evaluating object detection systems.

\begin{table}[!th]
\centering
\caption{COCO detection results on RetinaNet.}
\label{tab:coco_retinanet}
\setlength{\tabcolsep}{1.5mm}
\renewcommand{\arraystretch}{1.1}
\begin{tabular}{l|c|cccccc}
\toprule
Model       & Params (M) &  AP    & AP${}_{50}$  & AP${}_{75}$   & AP${}_{S}$    & AP${}_{M}$    & AP${}_{L}$    \\ \midrule
ResNet18 \cite{he2016deep} & 21.3  & 31.8 & 49.6 & 33.6 & 16.3 & 34.3 & 43.2 \\
PVTv1-Tiny \cite{wang2021pyramid} & 23.0  & 36.7 & 56.9 & 38.9 & 22.6 & 38.8 & 50.0 \\
EfficientViT-M4 \cite{liu2023efficientvit} & 8.8  & 32.7 & 52.2 & 34.1 & 17.6 & 35.3 & 46.0 \\
PVTv2-b0 \cite{wang2022pvt} & 13.0  & 37.2 & 57.2 & 39.5 & 23.1 & 40.4 & 49.7 \\
\rowcolor{mygray}
EfficientVMamba-T & 13.0  & 37.5 & 57.8 & 39.6 & 22.6 & 40.7 & 49.1 \\  \hline
ResNet50 \cite{he2016deep} & 37.7  & 36.3 & 55.3 & 38.6 & 19.3 & 40.0 & 48.8 \\
PVTv1-Small\cite{wang2021pyramid} & 34.2  & 40.4 & 61.3 & 43.0 & 25.0 & 42.9 & 55.7 \\
PVTv2-b1 \cite{wang2022pvt} & 23.8  & 41.2 & 61.9 & 43.9 & 25.4 & 44.5 & 54.3 \\
\rowcolor{mygray}
EfficientVMamba-S & 19.0  & 39.1 & 60.3 & 41.2 & 23.9 & 43.0 & 51.3 \\  \hline
ResNet101 \cite{he2016deep} & 56.7  & 38.5 & 57.8 & 41.2 & 21.4 & 42.6 & 51.1 \\
ResNeXt101-32x4d \cite{xie2017aggregated} & 56.4  & 39.9 & 59.6 & 42.7 & 22.3 & 44.2 & 52.5 \\
PVTv1-Medium \cite{wang2021pyramid} & 53.9  & 41.9 & 63.1 & 44.3 & 25.0 & 44.9 & 57.6 \\
\rowcolor{mygray}
EfficientVMamba-B & 44.0  & 42.8 & 63.9 & 45.8 & 27.3 & 46.9 & 55.1 \\  \bottomrule
\end{tabular}
\end{table}
\textbf{Results.} 
We summarize the results of RetinaNet detector in Table \ref{tab:coco_retinanet}. Remarkably, each variants competitively reducing the sizes while simultaneously exhibits a performance enhancement. 
The EfficientVMamba-T model stands out with 13M parameters and an AP of $37.5\%$, slightly higher by $5.7\%$ compared to the ResNet-18, which has $21.3M$ parameters. The performance of EfficientVMamba-T also surpasses PVTv1-Tiny by $0.8\%$ while matching it in terms of parameter count. EfficientVMamba-S, with only 19M parameters, achieves a commendable AP of $39.1\%$, outstripping the larger ResNet50 model, which shows a lower AP of $36.3\%$ despite having 37.7M parameters. In the higher echelons, EfficientVMamba-B, which boasts 44M parameters, secures an AP of $42.8\%$, signifying a significant lead over both ResNet101 and ResNeXt101-32x4d, highlighting the efficiency of our models even with a smaller parameter footprint. Notably, PVTv2-b0 with 13M parameters achieves an AP of $37.2\%$, which EfficientVMamba-T closely follows, indicating competitive performance with a similar parameter budget. For the comparisons with other backbones on Mask R-CNN, see Appendix.

\begin{table}[t]
    \centering
    \setlength{\tabcolsep}{1mm}
    \renewcommand{\arraystretch}{1.1}
    \caption{Results of semantic segmentation on ADE20K using UperNet \cite{xiao2018unified}. We measure the mIoU with single-scale (SS) and multi-scale (MS) testings on the \textit{val} set. The FLOPs are measured with an input size of $512\times2048$. MLN: multi-level neck.}
    \small
    \begin{tabular}{l|ccc|cc}
    \toprule
    Backbone & Image size & Params (M) & FLOPs (G) & mIoU (SS) & mIoU (MS)\\
    \midrule
    ResNet-50 \cite{he2016deep} & $512^2$ & 67 & 953 & 42.1 & 42.8\\
    ResNet-101 \cite{he2016deep} & $512^2$ & 85 & 1030 & 42.9 & 44.0\\
    Swin-T \cite{liu2021swin} & $512^2$ & 60 & 945 & 44.4 & 45.8\\
    Swin-S \cite{liu2021swin} & $512^2$ & 81 & 1039 & 47.6 & 49.5\\
    DeiT-S + MLN \cite{touvron2022deit} & $512^2$ & 58 & 1217 & 43.8 & 45.1\\
    DeiT-B + MLN \cite{touvron2022deit} & $512^2$ & 144 & 2007 & 45.5 & 47.2\\
    VMamba-T \cite{liu2024vmamba} & $512^2$ & 55 & 964 & 47.3 & 48.3\\
    VMamba-S \cite{liu2024vmamba} & $512^2$ & 76 & 1095 & 49.5 & 50.5\\
    \rowcolor{mygray}
    EfficientVMamba-T & $512^2$ & 14 & 230 & 38.9 & 39.3\\
    \rowcolor{mygray}
    EfficientVMamba-S & $512^2$ & 29 & 505 & 41.5 & 42.1\\
    \rowcolor{mygray}
    EfficientVMamba-B & $512^2$ & 65 & 930 & 46.5 & 47.3\\
    \bottomrule
    \end{tabular}
    \label{tab:ADE20K}
    \vspace{-1em}
\end{table}
\subsection{Semantic Segmentation} 
\label{sec:Semantic Segmentation}
\textbf{Training strategies.} Aligning with Vmamba \cite{liu2024vmamba} settings, we integrate an UperHead into the pre-trained model structure. Utilizing the AdamW optimizer, we initiate the learning rate at $6 \times 10^{-5}$. The fine-tuning stage consists of $160k$ iterations, using a batch size of 16. While the standard input resolution stands at $512 \times 512$, we also conduct experiments with $640 \times 640$ inputs and apply multi-scale (MS) testing to broaden our evaluation.

\textbf{Results.} The EfficientVMamba-T model yields mIoUs of 38.9\% (SS) and 39.3\% (MS), surpassing the ResNet-50's 42.1\% mIoU with far fewer parameters. EfficientVMamba-S achieves 41.5\% (SS) and 42.1\ (MS) mIoUs, bettering the DeiT-S + MLN despite having a lower computational footprint. The EfficientVMamba-B reaches 46.5\% (SS) and 47.3\% (MS), outperforming the heavier VMamba-S. These findings attest to the EfficientVMamba series’ balance of accuracy and computational efficiency in semantic segmentation.

\subsection{Ablation Study}
\label{sec:ablation}
\begin{table}[t]
\centering
\caption{Ablation Analysis: Evaluating the Efficacy of Enhanced Spatially Selective Dilatation (ES2D), Assessing the Synergistic Effect of Convolutional Branch Fusion Enhanced with Squeeze-and-Excitation (SE) Techniques, and Investigating the Role of Inverted Residual Blocks in Model Performance. For comparison with the baseline VMamba, we adjust the dimensions and number of layers of it to match the FLOPs.}\label{tab:ablation1}
\small
\renewcommand{\arraystretch}{1.1}
\setlength{\tabcolsep}{1.5mm}
\begin{tabular}{@{}l|ccc|ccc@{}}
\toprule
Model          & ES2D &  Fusion & InRes & Param (M) & FLOPs (G) & ACC (\%) \\ \midrule
VMamba-T & & & &2.4 & 0.9 & 72.1\\
\midrule
\multirow{3}{*}{EfficientVMamba-T} & \textcolor{green}{\cmark} & \textcolor{red}{\xmark}  & \textcolor{red}{\xmark} & 5         & 0.8      & 73.6     \\
                                   & \textcolor{green}{\cmark} & \textcolor{green}{\cmark}  & \textcolor{red}{\xmark} & 5         & 0.8      & 75.1     \\
                                   & \textcolor{green}{\cmark} & \textcolor{green}{\cmark}  & \textcolor{green}{\cmark} & 6         & 0.8      & 76.5     \\ \midrule
VMamba-B & & & & 16 & 4.2 & 80.2\\
\midrule
\multirow{3}{*}{EfficientVMamba-B} & \textcolor{green}{\cmark} & \textcolor{red}{\xmark}  & \textcolor{red}{\xmark} & 25        & 4.0      & 80.9     \\
                                   & \textcolor{green}{\cmark} & \textcolor{green}{\cmark}  & \textcolor{red}{\xmark} & 26        & 4.0      & 81.2     \\
                                   & \textcolor{green}{\cmark} & \textcolor{green}{\cmark}  & \textcolor{green}{\cmark} & 33        & 4.0      & 81.8     \\ \bottomrule
\end{tabular}
\vspace{-1em}
\end{table}

\begin{table}[t]
\centering
\caption{Comparisons of injecting convolution blocks at different stages on ImageNet dataset. We use EfficientVMamba-T in the experiments.}
\label{tab:ablation2}

\small
\renewcommand{\arraystretch}{1.1}
\setlength{\tabcolsep}{1.5mm}
\begin{tabular}{@{}l|cccc|ccc@{}}
\toprule
Layers   & Stage 1 & Stage 2 & Stage 3 & Stage 4 & Params (M) & FLOPs (G) & ACC (\%) \\ \midrule
EVSS only & EVSS & EVSS & EVSS & EVSS & 5 & 0.8 & 75.1\\
Previous & InRes   & InRes   & EVSS    & EVSS    & 5        & 0.8      & 75.6      \\
Ours     & EVSS    & EVSS    & InRes   & InRes   & 6        & 0.8     & 76.5      \\ \bottomrule
\end{tabular}
\vspace{-1.5em}
\end{table}
\textbf{Effect of atrous selective scan.}\label{sec:Effect of atrous selective scan} We implement experiment to validate the efficacy of atrous selective scan in Table \ref{tab:ablation1}. The upgrade from SS2D to ES2D significantly reduces the computational complexity from $0.8$ GFLOPs while retains competitive accuracy at $73.6\%$, a $1.5\%$ improvement on the tiny variant. Similarly, In the case of base variant, the model utilizing ES2D not only reduces the GFLOPs to $4.0$ from VMamba-B's $4.2$ but also exhibits an increase in accuracy from $80.2\%$ to $80.9\%$. The results suggest that the incorporation of ES2D in our EfficientVMamba models is one of the key factor in achieving the reduction of computational complexity by skip sampling while preserve the global respective field to keep competitive performance. The reduction of GLOPs also reveals the potency of ES2D in maintaining, and even enhancing, model accuracy while significantly reducing computational overhead, demonstrating its viability for resource-constrained scenarios.

\textbf{Effect of SSM-Conv fusion block.}
The integration of a convolutional branch following with a SE block enhances the performance of our model. For tiny variance, adding the local fusion feature extraction improves accuracy from 73.6\% to 75.1\%. In the case of EfficientVMamba-B, introducing fusion mechanism increases accuracy from 80.9\% to 81.2\%. The observed performance gains reveals the additional convolutional branch enhance the local feature extraction. By integrating Fusion, the models likely benefit from a more diversified feature set that captures a wider range of spatial details, improving the model's ability to generalize and thus boosting accuracy. This suggests that the strategic addition of such branches can effectively enhance the model's performance by providing a comprehensive and more nuanced respective field of the input feature map.

\textbf{Comparisons of injecting convolution block in different stages.}
In this paper, we obtain an interesting observation that our SSM based block, EVSS, is more beneficial in the early stages of the network. In contrast, previous works on light-weight ViTs usually inject the convolution blocks in the early stages and adopt Transformer blocks in the deep stages. As shown in Table \ref{tab:ablation2}, we compare the performance of injecting convolution blocks in different stages of EfficientVMamba-T, and the results indicate that, adopting Inverted Residual blocks in the deep stages with performs better than that in early stages. A explanation to the opposite phenomenons between our light-weight VSSMs and ViTs is that, the self-attention in Transformers has higher computation complexity and thus its computation at high resolutions is inefficient; while the SSMs, tailored for efficient modeling of long sequences, is more efficient and beneficial on capturing information globally at high resolutions.

\section{Conclusion}
This paper proposed EfficientVMamba, a lightweight state-space network architecture that adeptly combines the strengths of global and local information extraction, addressing the trade-off between model accuracy and computational efficiency. By incorporating an atrous-based selective scan with efficient skip sampling, EfficientVMamba ensures comprehensive global receptive field coverage while minimizing computational load. The integration of this scanning approach with a convolutional branch, followed by optimization through a Squeeze-and-Excitation module, allows for a robust re-balancing of global and local features. Additionally, the innovative use of inverted residual insertion further refines the model's multi-layer stages, enhancing its depth and effectiveness. Experimental results affirm that EfficientVMamba not only scales down computational complexity to $\mathcal{O}(N)$ but also delivers competitive performance across various vision tasks. The achievements of EfficientVMamba highlight its potential as a formidable framework in the evolution of lightweight, efficient, and general-purpose visual models.

%
%
\bibliographystyle{splncs04}
\bibliography{main}
\newpage
\section*{Appendix}
\subsection*{Comparisons with Other Backbones on Mask R-CNN.}
We also investigate the performance dynamics of our EfficientVMamba as a lightweight backbone within the Mask R-CNN schedule, as shown in Table \ref{tab:coco}. For the Mask R-CNN $1\times$ schedule, our EfficientVMamba-T model, with 11M parameters and $60G$ FLOPs, achieves an Average Precision (AP) of $35.6\%$. This is $1.6\%$ higher than ResNet-18, which has 31M parameters and 207G FLOPs. EfficientVMamba-S, with a greater number of parameters at 31M and 197G FLOPs, reaches an AP of $39.3\%$, which is $0.5\%$ above the ResNet-50 model with 44M parameters and 260G FLOPs. Our largest model, EfficientVMamba-B, shows a superior AP of 43.7\% with 53M parameters and a reduced computational requirement of 252G FLOPs, outperforming VMamba-T by $2.8\%$. In terms of Mask R-CNN $3\times$ MS schedule, EfficientVMamba-T maintains an AP of 38.3\%, surpassing ResNet-18's performance by $1.4\%$. The small variant records an AP of $41.5\%$, which is a $0.5\%$ improvement over PVT-T with a similar parameter count. Finally, EfficientVMamba-B achieves an AP of $45.0\%$, indicating a notable advancement of $2.2\%$ over VMamba-T.
\begin{table}[t]
    \centering
    \renewcommand{\arraystretch}{1.1}
    \setlength{\tabcolsep}{1.6mm}
    \caption{Object detection and instance segmentation results on COCO \textit{val} set.}
    \small
    \begin{tabular}{l|cc|ccc|ccc}
    \toprule
    \multicolumn{9}{c}{\textbf{Mask R-CNN 1$\times$ schedule}}\\
    \midrule
    Backbone & Params & FLOPs & AP$^\mathrm{b}$ & AP${}^\mathrm{b}_{50}$ & AP${}^\mathrm{b}_{75}$ & AP${}^\mathrm{m}$ & AP${}^\mathrm{m}_{50}$ & AP${}^\mathrm{m}_{75}$\\
    \midrule
    \rowcolor{mygray}
    EfficientVMamba-T & 11M & 60G & 35.6 & 57.7 & 38.0 & 33.2 & 54.4 & 35.1\\
    LightViT-T \cite{huang2022lightvit} & 28M & 187G & 37.8 & 60.7 & 40.4 & 35.9 & 57.8 & 38.0\\
    ResNet-18 \cite{he2016deep} & 31M & 207G & 34.0 & 54.0 & 36.7 & 31.2 & 51.0 & 32.7\\
    PVT-T \cite{wang2022pvt} & 33M & 208G & 36.7 & 59.2 & 39.3 & 35.1 & 56.7 & 37.3\\
    \rowcolor{mygray}
    EfficientVMamba-S & 31M & 197G & 39.3 & 61.8 & 42.6 & 36.7 & 58.9 & 39.2\\
    \midrule
    ResNet-50 \cite{he2016deep} & 44M & 260G & 38.2 & 58.8 & 41.4 & 34.7 & 55.7 & 37.2\\
    Swin-T \cite{liu2021swin} & 48M & 267G & 42.7 & 65.2 & 46.8 & 39.3 & 62.2 & 42.2 \\
    ConvNeXt-T \cite{liu2022convnet} & 48M & 262G & 44.2 & 66.6 & 48.3 & 40.1 & 63.3 & 42.8\\
    VMamba-T \cite{liu2024vmamba} & 42M & 286G & 46.5 & 68.5 & 50.7 & 42.1 & 65.5 & 45.3 \\
    \rowcolor{mygray}
    EfficientVMamba-B & 53M & 252G & 43.7 & 66.2 & 47.9 & 40.2 & 63.3 & 42.9\\
    \midrule
    \multicolumn{9}{c}{\textbf{Mask R-CNN 3$\times$ MS schedule}}\\
    \midrule
    \rowcolor{mygray}
    EfficientVMamba-T & 11M & 60G & 38.3 & 60.3 & 41.6 & 35.3 & 57.2 & 37.6\\
    ResNet-18 \cite{he2016deep} & 31M & 207G & 36.9 & 57.1 & 40.0 & 33.6 & 53.9 & 35.7\\
    PVT-T \cite{wang2022pvt} & 33M & 208G & 39.8 & 62.2 & 43.0 & 37.4 & 59.3 & 39.9\\ 
    LightViT-T \cite{huang2022lightvit} & 28M & 187G & 41.5 & 64.4 & 45.1 & 38.4 & 61.2 & 40.8\\
    \rowcolor{mygray}
    EfficientVMamba-S & 31M & 197G & 41.6 & 63.9 & 45.6 & 38.2 & 60.8 & 40.7\\
    \midrule
    ResNet-50 \cite{he2016deep} & 44M & 260G & 41.0 & 61.7 & 44.9 & 37.1 & 58.4 & 40.1\\
    Swin-T \cite{liu2021swin} & 48M & 267G & 46.0 & 68.1 & 50.3 & 41.6 & 65.1 & 44.9 \\
    ConvNeXt-T \cite{liu2022convnet} & 48M & 262G & 46.2 & 67.9 & 50.8 & 41.7 & 65.0 & 44.9 \\
    VMamba-T \cite{liu2024vmamba} & 42M & 286G & 48.5 & 69.9 & 52.9 & 43.2 & 66.8 & 46.3 \\
    \rowcolor{mygray}
    EfficientVMamba-B & 53M & 252G & 45.0 & 66.9 & 49.2 & 40.8 & 64.1 & 43.7\\
    \bottomrule
    \end{tabular}
    \label{tab:coco}
    \vspace{-1.5em}
\end{table}

\subsection*{Comparisons with MobileNetV2 Backbone}
We compare variant architectures and reveal a significant performance difference based on the integration of our innovative block, EVSS, versus Inverted Residual (InRes) blocks at specific stages.
\begin{table}[ht]
\centering
\caption{Comparisons of MobileNetV2 (All stages composed with EVSS.) on ImageNet dataset. We assess both tiny and base models on the ImageNet.} \label{tab:appendix1}
\setlength{\tabcolsep}{1mm}
\renewcommand{\arraystretch}{1.1}
\begin{tabular}{l|cccc|ccc}
\hline
Variants & \multicolumn{1}{l}{Stage 1} & Stage 2 & Stage 3 & Stage 4 & Params (M) & FLOPs (G) & ACC (\%) \\ \hline
\multirow{3}{*}{Tiny} & InRes & InRes & InRes & InRes &  6  &   0.7     & 76.0 \\
                      & EVSS  & EVSS  & EVSS  & EVSS  &  5  &   0.8  & 75.1 \\
                      \rowcolor{mygray}
                      & EVSS  & EVSS  & InRes & InRes & 6  & 0.8 & 76.5 \\ \hline
\multirow{3}{*}{Base} & InRes & InRes & InRes & InRes &  34  &  3.8   & 81.4 \\
                      & EVSS  & EVSS  & EVSS  & EVSS  &  26  &  4.0   & 81.2 \\
                      \rowcolor{mygray}
                      & EVSS  & EVSS  & InRes & InRes & 33 & 4.0 & 81.8 \\ \hline
\end{tabular}
\end{table}
This results in Table \ref{tab:appendix1} shows that using InRes consistently across all stages in both tiny and base variants achieves a good performance, with the base variant notably reaching an accuracy of 81.4\%. When EVSS is applied across all stages (the strategy of MobileNetV2 \cite{sandler2018mobilenetv2}), we observe a slight decrease in accuracy for both variants, suggesting a nuanced balance between architectural consistency and computational efficiency. Our fusion approach that combines EVSS in the initial stages with InRes in the later stages enhances accuracy to 76.5\% and 81.8\% for the tiny and base variants, respectively. This strategy benefits from the early-stage efficiency of EVSS and the advanced-stage convolutional capabilities of InRes, thus optimizing network performance by leveraging the strengths of both block types with limit computational resources.

\subsection*{Limitations}
Visual state space models that operate with a linear-time complexity $\mathcal{O}(N)$ relative to sequence length demonstrate marked enhancements, particularly in high-resolution downstream tasks, which contrasted with prior CNN-based and Transformer-based models.
However, the computational design of SSMs inherently exhibits increased computational sophistication than both convolutional and self-attention mechanisms, which complicates the performance of efficient parallel processing. There remains promising potential for future investigation on optimizing the computational efficiency and scalability of visual state space models (SSMs). 

\end{document}


\maketitle
\section{Appendix}

\subsection{Comparisons with Other Light-weight Backbones on RetinaNet}

We also investigate the performance dynamics of our EfficientVMamba as a lightweight backbone within the RetinaNet framework. 
\begin{table}[th]
\centering
\caption{COCO detection results on RetinaNet.}
\setlength{\tabcolsep}{1.5mm}
\renewcommand{\arraystretch}{1.1}
\begin{tabular}{l|c|cccccc}
\toprule
Model       & Params (M) &  AP    & AP${}_{50}$  & AP${}_{75}$   & AP${}_{S}$    & AP${}_{M}$    & AP${}_{L}$    \\ \midrule
ResNet18 \cite{he2016deep} & 21.3  & 31.8 & 49.6 & 33.6 & 16.3 & 34.3 & 43.2 \\
PVTv1-Tiny \cite{wang2021pyramid} & 23.0  & 36.7 & 56.9 & 38.9 & 22.6 & 38.8 & 50.0 \\
EfficientViT-M4 \cite{liu2023efficientvit} & 8.8  & 32.7 & 52.2 & 34.1 & 17.6 & 35.3 & 46.0 \\
PVTv2-b0 \cite{wang2022pvt} & 13.0  & 37.2 & 57.2 & 39.5 & 23.1 & 40.4 & 49.7 \\
\rowcolor{mygray}
EfficientVMamba-T & 13.0  & 37.5 & 57.8 & 39.6 & 22.6 & 40.7 & 49.1 \\  \hline
ResNet50 \cite{he2016deep} & 37.7  & 36.3 & 55.3 & 38.6 & 19.3 & 40.0 & 48.8 \\
PVTv1-Small\cite{wang2021pyramid} & 34.2  & 40.4 & 61.3 & 43.0 & 25.0 & 42.9 & 55.7 \\
PVTv2-b1 \cite{wang2022pvt} & 23.8  & 41.2 & 61.9 & 43.9 & 25.4 & 44.5 & 54.3 \\
\rowcolor{mygray}
EfficientVMamba-S & 19.0  & 39.1 & 60.3 & 41.2 & 23.9 & 43.0 & 51.3 \\  \hline
ResNet101 \cite{he2016deep} & 56.7  & 38.5 & 57.8 & 41.2 & 21.4 & 42.6 & 51.1 \\
ResNeXt101-32x4d \cite{xie2017aggregated} & 56.4  & 39.9 & 59.6 & 42.7 & 22.3 & 44.2 & 52.5 \\
PVTv1-Medium \cite{wang2021pyramid} & 53.9  & 41.9 & 63.1 & 44.3 & 25.0 & 44.9 & 57.6 \\
\rowcolor{mygray}
EfficientVMamba-B & 44.0  & 42.8 & 63.9 & 45.8 & 27.3 & 46.9 & 55.1 \\  \bottomrule
\end{tabular}
\end{table}
Remarkably, each variants competitively reducing the sizes while simultaneously exhibits a performance enhancement. For instance, EfficientVMamba-T outperforms PVTv1-Tiny by registering improvements of ($+0.8\%/+0.9\%$) on $AP$ and $AP_{50}$,  while simultaneously reducing the parameter count from $23$M to $13$M respectively. Similarly, EfficientVMamba-B surpasses PVTv1-Medium with enhancements of (+0.9\%/+0.8\%) on $AP$ and $AP_{50}$, and decreases the parameter count from $53.9$M to $44$M.
The advancements are particularly impressive as our model underscores the ability to achieve higher detection accuracy without the typical trade-off in size, showing its potential for applications where model compactness and efficiency are paramount.






\subsection{Comparisons with MobileNetV2 Backbone}
We compare variant architectures and reveal a significant performance difference based on the integration of our innovative block, EVSS, versus Inverted Residual (InRes) blocks at specific stages.
\begin{table}[ht]
\centering
\caption{Comparisons of MobileNetV2 (All stages composed with EVSS.) on ImageNet dataset. We assess both tiny and base models on the ImageNet.} \label{tab:appendix1}
\setlength{\tabcolsep}{1mm}
\renewcommand{\arraystretch}{1.1}
\begin{tabular}{l|cccc|ccc}
\hline
Variants & \multicolumn{1}{l}{Stage 1} & Stage 2 & Stage 3 & Stage 4 & Params (M) & FLOPs (G) & ACC (\%) \\ \hline
\multirow{3}{*}{Tiny} & InRes & InRes & InRes & InRes &  6  &   0.7     & 76.0 \\
                      & EVSS  & EVSS  & EVSS  & EVSS  &  5  &   0.8  & 75.1 \\
                      \rowcolor{mygray}
                      & EVSS  & EVSS  & InRes & InRes & 6  & 0.8 & 76.5 \\ \hline
\multirow{3}{*}{Base} & InRes & InRes & InRes & InRes &  34  &  3.8   & 81.4 \\
                      & EVSS  & EVSS  & EVSS  & EVSS  &  26  &  4.0   & 81.2 \\
                      \rowcolor{mygray}
                      & EVSS  & EVSS  & InRes & InRes & 33 & 4.0 & 81.8 \\ \hline
\end{tabular}
\end{table}
This results in Table \ref{tab:appendix1} shows that using InRes consistently across all stages in both tiny and base variants achieves a good performance, with the base variant notably reaching an accuracy of 81.4\%. When EVSS is applied across all stages (the strategy of MobileNetV2 \cite{sandler2018mobilenetv2}), we observe a slight decrease in accuracy for both variants, suggesting a nuanced balance between architectural consistency and computational efficiency. Our fusion approach that combines EVSS in the initial stages with InRes in the later stages enhances accuracy to 76.5\% and 81.8\% for the tiny and base variants, respectively. This strategy benefits from the early-stage efficiency of EVSS and the advanced-stage convolutional capabilities of InRes, thus optimizing network performance by leveraging the strengths of both block types with limit computational resources.






\subsection{Limitations}
Visual state space models that operate with a linear-time complexity $\mathcal{O}(N)$ relative to sequence length demonstrate marked enhancements, particularly in high-resolution downstream tasks, which contrasted with prior CNN-based and Transformer-based models.
However, the computational design of SSMs inherently exhibits increased computational sophistication than both convolutional and self-attention mechanisms, which complicates the performance of efficient parallel processing. There remains promising potential for future investigation on optimizing the computational efficiency and scalability of visual state space models (SSMs). 
%
%
\bibliographystyle{splncs04}
\bibliography{main}